\documentclass{article}

\usepackage{microtype}
\usepackage{graphicx}
\usepackage{subfigure}
\usepackage{booktabs} % for professional tables
\usepackage{amsfonts}

\usepackage{hyperref}

\usepackage{tabularx}
\usepackage{spverbatim}
\usepackage{listings}
\usepackage{natbib}
\usepackage{amsmath}
\usepackage[accepted]{icml2021}

\icmltitlerunning{Domain Adversarial Reinforcement Learning}

\begin{document}

% \section*{To Do's}
% \input{to_do_s}
% \clearpage

\twocolumn[
\icmltitle{Domain Adversarial Reinforcement Learning}



\icmlsetsymbol{equal}{*}

\begin{icmlauthorlist}
% \icmlauthor{Aeiau Zzzz}{equal,to}
% \icmlauthor{Bauiu C.~Yyyy}{equal,to,goo}
\icmlauthor{Bonnie Li}{mc,mi}
\icmlauthor{Vincent Fran\c cois-Lavet}{vu}
\icmlauthor{Thang Doan}{mc,mi}
\icmlauthor{Joelle Pineau}{mc,mi}
\end{icmlauthorlist}

\icmlaffiliation{mc}{McGill University}
\icmlaffiliation{mi}{Mila}
\icmlaffiliation{vu}{VU Amsterdam}
% \icmlaffiliation{ed}{School of Computation, University of Edenborrow, Edenborrow, United Kingdom}

\icmlcorrespondingauthor{Bonnie Li}{bonnie.li@mail.mcgill.ca}
% \icmlcorrespondingauthor{Eee Pppp}{ep@eden.co.uk}

\icmlkeywords{Machine Learning, ICML}

\vskip 0.3in
]



\printAffiliationsAndNotice{}  % leave blank if no need to mention equal contribution

\begin{abstract}
We consider the problem of generalization in reinforcement learning where visual aspects of the observations might differ, e.g. when there are different backgrounds or change in contrast, brightness, etc. We assume that our agent has access to only a few of the MDPs from the MDP distribution during training. The performance of the agent is then reported on new unknown test domains drawn from the distribution (e.g. unseen backgrounds). For this ``zero-shot RL" task, we enforce invariance of the learned representations to visual domains via a domain adversarial optimization process. We empirically show that this approach allows achieving a significant generalization improvement to new unseen domains.
\end{abstract}

\section{Introduction}
\label{submission}

\begin{figure}[ht!]
\begin{center}
\includegraphics[width=0.46\textwidth]{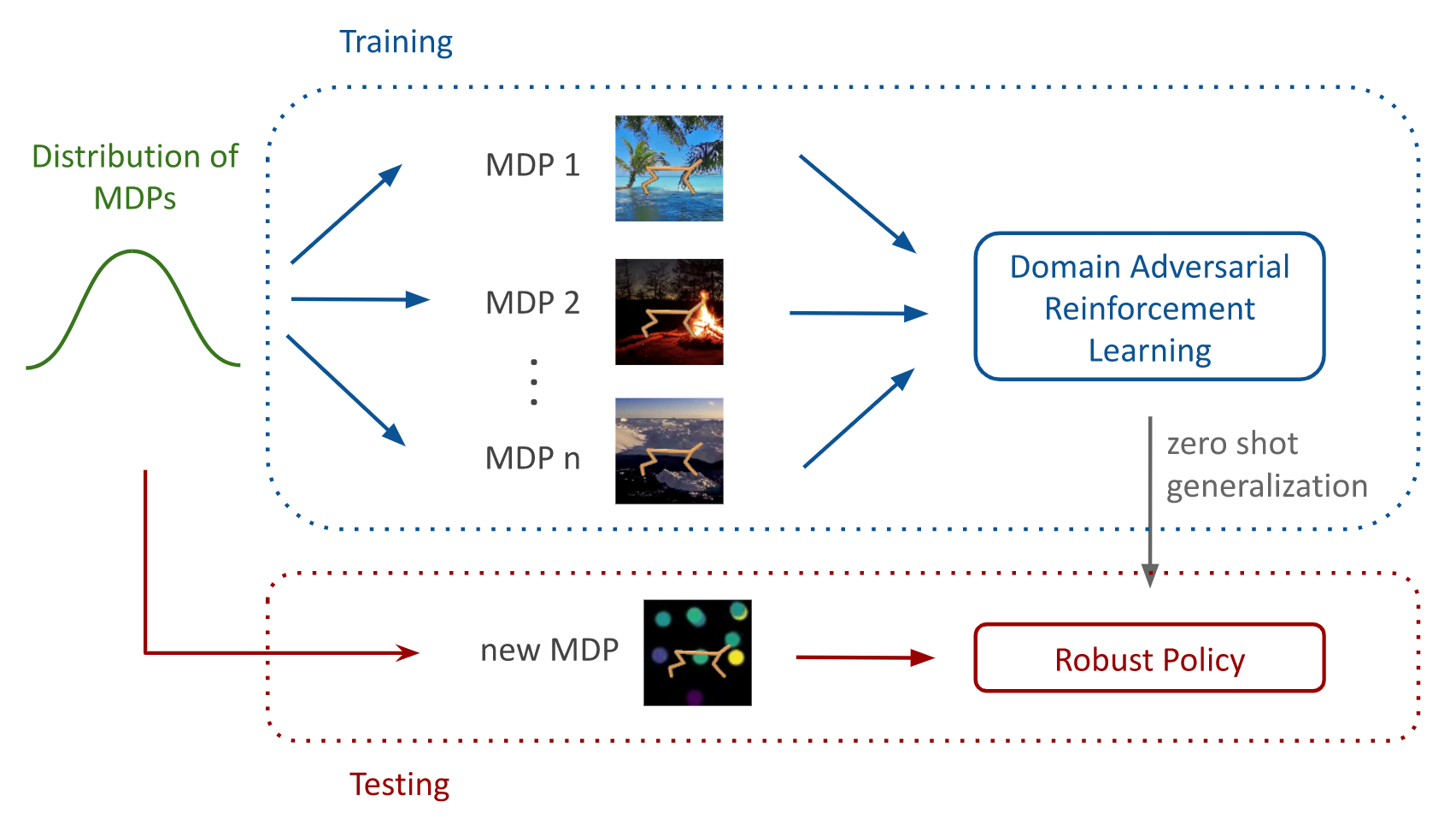}
\end{center} 
\caption{% Setup of our method
Training and evaluation set up: the agent is trained in a distribution of MDPs with different visual backgrounds and evaluation is done in new domains with unknown backgrounds. To improve generalization to new MDPs from the distribution, our domain adversarial approach is specifically trained to focus on the important visual aspects of the tasks and ignore the irrelevant factors.
}
\label{set_up}
\end{figure}

Deep reinforcement learning (deep RL) has achieved great success in solving complex control tasks from high-dimensional pixel inputs, such as video games \citep{ale,procgen}. 
However, real world reinforcement learning remains a challenge. For applications such as robotics, it can be expensive and dangerous (no guarantee on the policy safety). As such, simulation is often used as a cheaper and safer alternative \citep{chaffre2020sim,matas2018sim}.
But RL agents can overfit very easily to the environment in which they were trained \citep{cobbe2019quantifying,zhang2018study}. Even for the same task dynamics, RL agents can be prone to brittleness caused by different colors, noise injection in the observation space, distractors, or background change \citep{ding2020mutual, natural_env, dbc}.
As illustrated in Figure \ref{set_up}, we consider the setting where the agent is trained on a limited number of environments that share the same underlying task dynamics but where the observation distributions might differ.

Learning a robust representation is key towards generalization in deep RL. The representation should retain sufficient task-relevant information in order to learn a good policy, while ignoring irrelevant features of the environment. Reconstruction-based methods \citep{ha2018world,sac_ae} provide a good approach for compressing the representation but do not remove irrelevant features for the task, which may lead to overfitting. 
% That representation, capturing either spatial \citep{curl} or temporal \citep{mazoure2020deep} generative factors, should allow the RL agent to learn a set of \emp{invariant} features across different visual domains, such as changes in background).
Other approaches aim at constructing representations that go beyond pixels similarity by capturing %spatial, temporal generative factors, or bisimulation-
relevant factors in the dynamics to increase sample efficiency of the agent \citep{franccois2019combined, curl, mazoure2020deep, dbc}.%, and features invariant to visual changes in the observation. 

%This problem can be seen as a POMDP with an observable context that influences the observation given the current underlying state. %This has been formlaized as a block MDP\cite{du2019provably}

% Previous works have tackled this problem with specific transfer learning techniques that is able to adapt a policy from one source domain to a target domain (see Section \ref{sec:related_work}). %\cite{gamrian2019transfer, wappo}. 

In this work, we propose a specific approach that improves generalization in the context of tasks that share the same dynamics but different visual domains. Our approach aims to learn an abstract representation space at the output of an encoder that (i) captures the important features from the source domains and (ii) is invariant with respect to the domain distribution shift. We evaluate our approach on DeepMind Control tasks, and examine zero-shot generalization to unseen visual domains with both stationary and non-stationary backgrounds.

\section{Related Works}
\subsection{Feature Distribution Alignment}
% Unsupervised domain adaption is the setting where labeled training data is available on a source domain, but the goal is to have good performance on a target domain with only unlabeled data. 
% The main approach is to align the source and the target domains in some feature space. This is done by optimizing some measure of discrepancy. 
Our paper builds on early works for feature distribution alignment. Aligning distributions in the feature space can be done by optimizing some measure of discrepancy, such as the maximum mean discrepancy (MMD) \cite{pan2010domain,long2017deep}. 

Another popular approach for aligning distributions is via an adversarial discriminator that distinguishes between the feature distributions. This was used in the setting of domain adaptation for supervised learning, where labeled training data is available on a source domain, but the goal is to have good performance on a target domain with only unlabeled data. By minimizing the domain classification accuracy of the adversarial discriminator, one can align the feature representations across source domain and target domain \cite{ajakan2014domain,dann,adda}.

We extend this approach to the RL context 
while having access to only a few training domains 
and provide experimental results that show how the features learned in this setting can zero-shot transfer to unseen domains.

\subsection{Visual Transfer in Reinforcement Learning}
Reinforcement learning environments with different visual backgrounds can be formalized as a block MDP. In this setting, new (unseen) tasks may
appear to the agent as just another variation \cite{parisotto2015actor}, however there must be enough training environments.

%cited in the preliminary section: (\citet{latent_decoding}, \citet{misa}). 
Several recent works examined various instances of generalization and transfer within these environments.
Robust Domain Randomization (RDR, \citet{rdr}) considers the problem of visual randomization, and attempts to zero-shot transfer to new domain by minimizing the Euclidean distance between feature representations across randomization of training domains.
MISA (\citet{misa}) uses tools from causal inference, namely ICP and IRM, to learn Model-Irrelevance State Abstractions, which shows generalization on low-dimensional RL tasks and visual imitation learning.
PAD (\citet{pad}) employs a self-supervised inverse dynamic model combined with data augmentation for faster adaptation during deployment when the policy is deployed in a different domain.

One close work to ours is WAPPO (\citet{wappo}) that investigates the problem of unsupervised domain adaptation in the RL setting, and uses Wasserstein GAN to measure and align the feature distributions from the source and target domain.
% Our work is most closely related to WAPPO, but we consider the "zero-shot RL" setting for which we aim at generalizing to new unseen domains from a distribution of training environments.
A major difference is that WAPPO trains on one single source domain and requires access to the target domain observations during training. Our setting directly evaluate on unseen domains (zero-shot) without any prior knowledge on the distribution of the environments.

% While WAPPO considers one source and one target domain, and requires access to target domain observations during training, in our setting we aim to zero-shot generalize to unseen domains from training on a distribution of environments.

\section{Preliminaries/Background}
\subsection{Reinforcement learning}

%In this paper we follow the definition of block Markov decision process given by \citet{du2019provably}. The environment is described by a finite, but unobservable latent state space $\mathcal{S}$, a finite action space $\mathcal{A}$, and a context space X. The dynamics of a BMDP is described by the initial state $s_0 \in \mathcal S$ and two conditional probability functions: the state-transition function p and context-emission function q, defining conditional probabili- ties $\mathbb(s′|s,a)$ and $q(x|s), \forall s,s′ \in \mathcal S, a \in \mathcal {A}, x \in \mathcal {X}$.

The standard Markov Decision Process (MDP) is formalized as 
$\mathcal{M}:=
\left\langle
    \mathcal{S},
    \mathcal{A},
    r,
    P,
    S_0,
    \gamma
\right\rangle
$,
where 
$\mathcal{S}$ is the state space;
$\mathcal{A}$ is the action space; given states $s$, $s'\in\mathcal{S}$, action $a\in\mathcal{A}$,
$P(s'|s, a)$ is the transition probability of transition from $s$ to $s'$ under the action $a$ and
$r(s, a)$ is the reward collected at state $s$ after executing the action $a$;
$S_0$ is the initial distribution of states and
$\gamma$ is the discount factor.

At each discrete time steps $t \in \mathbb{N}$, the agent executes an action $a_t$ in its environment according to some policies. We define a stochastic policy $\pi: \mathcal{S}~\rightarrow~\mathbb P(\mathcal{A})$ such that $\pi(a|s)$ is the conditional probability that the agent takes action $a\in\mathcal{A}$ after observing the state $s\in\mathcal{S}$.

%\begin{align}
%    \eta(\pi)=\mathbb{E}_{\pi}\left[\sum_{t=0}^\infty\gamma^tr(s_t, a_t)\right]
%    \label{eq:eta_mdp}
%\end{align}
We define the state value function $V_\pi$ and the state-action value function $Q_\pi$ as:
\begin{align*}
    V_{\pi}(s)&=\mathbb{E}_{\pi}\left[\sum_{k=t}^\infty\gamma^{t-k}r(s_k, a_k)\bigg|s_t=s\right],\\
    Q_{\pi}(s, a)&=\mathbb{E}_{\pi}\left[\sum_{k=t}^\infty\gamma^{t-k}r(s_k, a_k)\bigg|s_t=s,a_t=a\right].
\end{align*}
The objective is to find a policy $\pi$ which maximizes the expected discounted reward $V_{\pi}$.
The difference between $Q_\pi$ and $V_\pi$ is known as the advantage function:
 \begin{align*}
     A_{\pi}(s,a)&=Q_{\pi}(s, a)-V_{\pi}(s).
 \end{align*}
% where $a \sim \pi(\cdot|s)$ and $s' \sim P(\cdot|s,a)$. 
% The coverage of policy $\pi$ is defined as the stationary state visitation probability $\rho_\pi(s)=\mathbb{P}(s_t=s)$ under $\pi$. 
 
\subsection{Block MDPs}
Our problem set up of reinforcement learning environments with different visual backgrounds can be formalized as a block MDP, defined by $\mathcal{M}:=
\left\langle
    \mathcal{S},
    \mathcal{A},
    \mathcal{X},
    p,
    q,
    R,
    \gamma
\right\rangle
$. 
% For related block MDPs drawn from such block 
Such block MDP defines a distributions of MDPs. 
This setting assumes that the related MDPs would share the same unobserved state space $\mathcal{S}$, 
action space $\mathcal{A}$, 
latent transition distribution $p(s'\mid s, a)$, 
and reward function $R$. 
But the MDPs can differ in their observed state space $\mathcal{X}$ (also formalized as the context space \citep{latent_decoding}), 
and emission function $q(x\mid s)$, 
which maps the hidden state space to the observed context space.
% \bonnie{add the block mdp assumptions, and settings}

We make a few relaxations and assumptions from the original block structure assumption \citep{latent_decoding} for our problem set up. 
Similarly to \citet{misa}, 
we assume that there exists an invariant state embedding from the observed state space: $\mathcal{S} \mapsto \mathcal{Z}$. 
We further assume the existance of an invariant dynamic model and reward model within such state embedding $z \in \mathcal{Z}$. 
When considering the policy applied from the state embedding, this implies that optimal behaviour is achievable with the same policy across all MDPs within the block.
 
\subsection{Soft Actor Critic}
\label{sac}
Soft Actor Critic (SAC, \citet{sac}) is an off-policy actor critic method using the maximum entropy framework. It trains a neural network $Q_\theta$ to approximate the agent's current Q-function, by minimizing the soft Bellman residual
$$L_Q = \Big(Q(s_t, a_t) - \big(r_t + \gamma \Bar{V}(s_{t+1}\big)\Big)^2 $$
where $\Bar{V}$ is the target value function approximated as
$$\Bar{V}(s_t) = \mathbb E_{a_t \sim \pi} \left[ \Bar{Q}(s_t, a_t) - \alpha \ln \pi(a_t \mid s_t) \right]$$
where $\Bar{Q}$ is the target Q function, computed as the exponential moving average of $Q_\theta$. The policy is trained by minimizing the divergence between the policy and the exponential of the soft Q function
% between the policy and a Bolzmann distribution induced by the Q function
$$L_\pi = D_{KL} \Big(\pi(\cdot \mid s_t) \parallel \exp(\frac{1}{\alpha}Q(s_t, \cdot)) \Big) $$

Several recent works (\citet{sac_ae}, \citet{curl}, \citet{rad}, \citet{drq}) use representation loss and data augmentation to improve SAC from pixels. Data augmentation, namely random crop alone, was shown to be particularly effective. Following RAD \citet{rad}, we build on SAC with randomly cropped input pixel observations.

% ARCHITECTURE ILLUSTRATION
\begin{figure*}[t]
\includegraphics[width=1\textwidth]{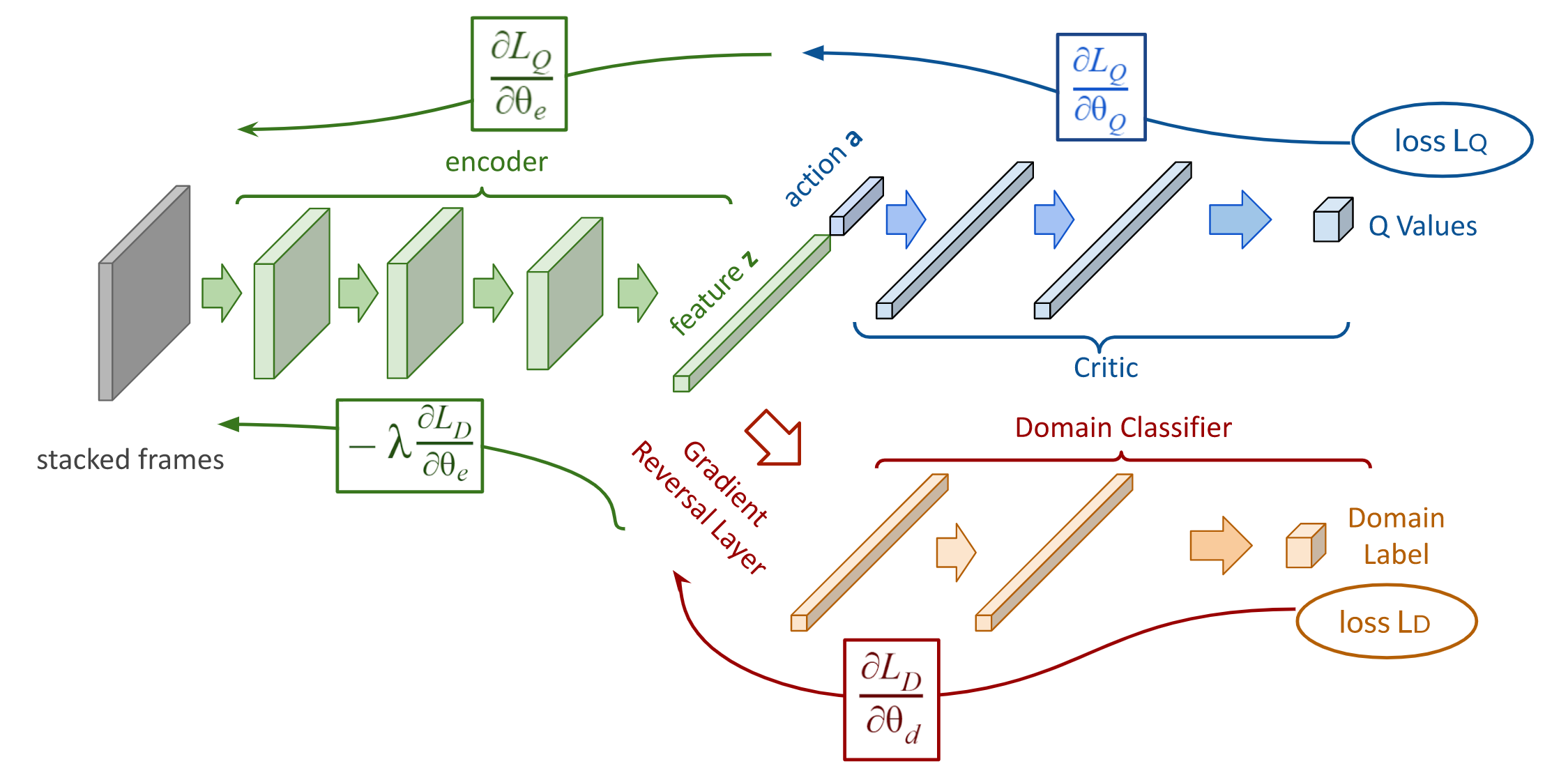}
\caption{The proposed DARL architecture builds a domain classification module (red) along with the critic and encoder of SAC.
Feature distributions across domains are aligned via the gradient reversal layer between the encoder and the domain classifier. 
The gradient reversal layer multiplies the gradient by a certain negative constant during back propagation. 
Otherwise, the training follows the standard procedure and minimizes the critic loss and domain classification loss. 
Gradient reversal ensures that features across domains are indistinguishable for the domain classifier, which encourages domain invariant features.
}\label{architecture}
\end{figure*}
\section{DARL: Domain Adversarial Reinforcement Learning}
% ARCHITECTURE ILLUSTRATION
% \begin{figure*}[t]
% \includegraphics[width=1\textwidth]{img/architecture.png}
% \caption{The proposed architecture builds a domain classification module (red) along with the critic and encoder of SAC.
% Feature distributions across domains are aligned via the gradient reversal layer between the encoder and the domain classifier. 
% The gradient reversal layer multiplies the gradient by a certain negative constant during back propagation. 
% Otherwise, the training follows the standard procedure and minimizes the critic loss and domain classification loss. 
% Gradient reversal ensures that features across domains are indistinguisable for the domain classifier, which encourages domain invariant features.
% }\label{architecture}
% \end{figure*}

We train an agent to maximize its expected cumulative reward over a distribution of MDPs defined by a block MDP.
The agent should not only maximize rewards on the training environments, but should also generalize well to unseen environments drawn from the same block without any additional training.
To do so, we propose to combine a strong model-free RL algorithm, SAC (with data augmentation), with a proposed auxillary adversarial loss to improve generalization. 

% \subsection{Domain Adversarial Reinforcement Learning}
We aim to learn a robust representations from high-dimensional observations that can directly generalize to 
diverse MDP contexts drawn from the same block,
i.e. with the same underlying dynamics but different visual settings.
Here a good representation would be one that is informative enough for good task performance, and is general enough to generalize zero-shot to unseen visual settings.

Intuitively, if we learn representations that are invariant to background information, the representations can then generalize to unseen backgrounds. 
To explicitly remove the backgrounds, we propose to use an adversarial discriminator to measure and align feature distributions across domains. 
We use the adversarial discriminator to predict the 
% background label of the observations, 
context of the MDP (i.e. label of the background in our setting), 
and try to force the features to be as indistinguishable as possible for the discriminator via a gradient reversal layer. See Figure \ref{architecture} for the proposed architecture. We now describe our approach in detail. 

% ARCHITECTURE ILLUSTRATION
% \begin{figure*}[t]
% \includegraphics[width=1\textwidth]{img/architecture.png}
% \caption{The proposed architecture builds a domain classification module (red) along with the critic and encoder of SAC.
% Feature distributions across domains are aligned via the gradient reversal layer between the encoder and the domain classifier. 
% The gradient reversal layer multiplies the gradient by a certain negative constant during back propagation. 
% Otherwise, the training follows the standard procedure and minimizes the critic loss and domain classification loss. 
% Gradient reversal ensures that features across domains are indistinguisable for the domain classifier, which encourages domain invariant features.
% }\label{architecture}
% \end{figure*}

% Joelle: first describe main components
% encoder, critic
% \textbf{Actor Critic and Encoder. }
We use an encoder $E_\theta: \mathcal{S} \mapsto \mathcal{Z}$, parameterized by $\theta_e$, to map the high dimensional states into lower-dimensional representations. 
We use the resulting feature vectors $z$ for the policy $\pi_\theta$ and the critic $Q_\theta$, parametrized by $\theta_\pi$ and $\theta_Q$ respectively. 
Building on SAC, the policy is trained with the encoder detached, i.e. 
% we detach the feature vector when used as input to the actor. 
the policy loss will not backpropagate to the encoder.
% We concatenate the state features with actions as inputs to the critic. 
The encoder is trained jointly with the critic, for which we concatenate the state features with actions as inputs.
We have
$$
\forall s \in \mathcal{S},\;\;
z = E_\theta(s;\theta_e) ,\; 
$$
$$
Q(s, a) \approx Q_\theta(z, a; \theta_Q) ,\;
a \sim \pi_\theta(a \mid \Bar{z}) ,\;
$$
where 
$s$ is the high dimensional state, 
$z$ is the encoded state representation, 
and $\Bar{z}$ denotes the state representation detached from encoder.

\textbf{Adversarial Discriminator.}
The adversarial discriminator, in our setting, is the domain classifier $D_\theta$ predicting the domain label (background) from the state. This is trained with the domain classification loss $L_D(s, y)$, which is the negative log-likelihood of the correct domain label output
$$L_D(s, y) = - \ln P_D(y \mid z), z = E_\theta(s) \;,$$
% with a softmax activation function 
where $P_D$ is the discriminator output of the domain probabilities, which are normalized via a softmax activation function, 
and $y$ is the correct domain label of state $s$.

\textbf{Gradient Reversal Layer. }
Following Domain Adversarial Neural Network (DANN, \citet{dann}), we connect the adversarial discriminator with the encoder via a gradient reversal layer (GRL). 
This layer reverses the gradient during backpropapagtion, encouraging the encoder the learn against the discriminator.
Mathematically, we can formalize the gradient reversal layer as a "pseudo function" $G_{\lambda}$ characterized by the following two (incompatible) equations:
$$ G_{\lambda} (z) = z \;,\; \frac{d G_{\lambda}}{dz} = - \lambda I \;,$$
where $I$ is the identity matrix.
During the forward pass, GRL is the identity transform. 
During backpropagation, GRL takes the gradient from the subsequent level, multiplies it by a pre-defined negative constant $-\lambda$ and passes it to the preceding layer. 

\textbf{Adaption factor $\lambda$. } 
% \bonnie{add about adaption factor and scaling parameter}
The gradient reversal layer has only one hyper-parameter $\lambda$, which is a constant that scales the gradient during backpropagation. We use this constant to mitigate the effect of noise from the discriminator in the early training stage, by increasing $\lambda$ from 0 to 1 during the course of training. Similarly to \citet{dann}, we update $\lambda$ using the following
$$\lambda_p = \frac{2}{1 + \exp(-10 \cdot p)} -1 \;,$$
where $p$ is increased linearly from 0 to 1 over the first million environment steps of training, i.e. $p = \text{min}(\frac{\text{current \ step}}{10^6}$, 1).

% \textbf{Adversarial Discriminator.}
% The adversarial discriminator, in our setting, is the domain classifier predicting the domain label of the state from state features. This is trained with the domain classification loss $L_D(s, y)$, which is the negative log-likelihood of the correct domain label output.
% $$L_D(s, y) = - \ln P_D(y \mid z), z = E_\theta(s) \;,$$
% where $z$ is the features of state $s$ from the encoder $E_\theta$.

With the gradient reversal layer between encoder features and domain classifier, by directly minimizing the domain classification loss via backpropagation, we train the domain classifier $D_\theta$ to distinguish the background class from state features, and simultaneously train the encoder $E_\theta$ to produce state features that \textit{maximizes} the classification error (by taking the reversed gradient), therefore encouraging invariant features that are indistinguishable across domains.

Overall, we jointly minimize the critic loss and the domain classification loss. 
Separately, the policy is trained (with the encoder detached) with policy loss $L\pi$ following from SAC in \ref{sac}.
% the encoder is trained jointly with the critic and domain classifier. 
To jointly train the encoder, critic, and domain classifier, we minimize
$$L_\theta = L_Q + \beta \cdot L_D \;,$$
where $L_D$ is the discriminator loss described above, 
$L_Q$ follows from the Q loss in SAC in \ref{sac}, and $\beta$ scales the two losses.

\section{Experiments}
\begin{figure*}[ht!]
\begin{center}
\includegraphics[width=0.7\textwidth]{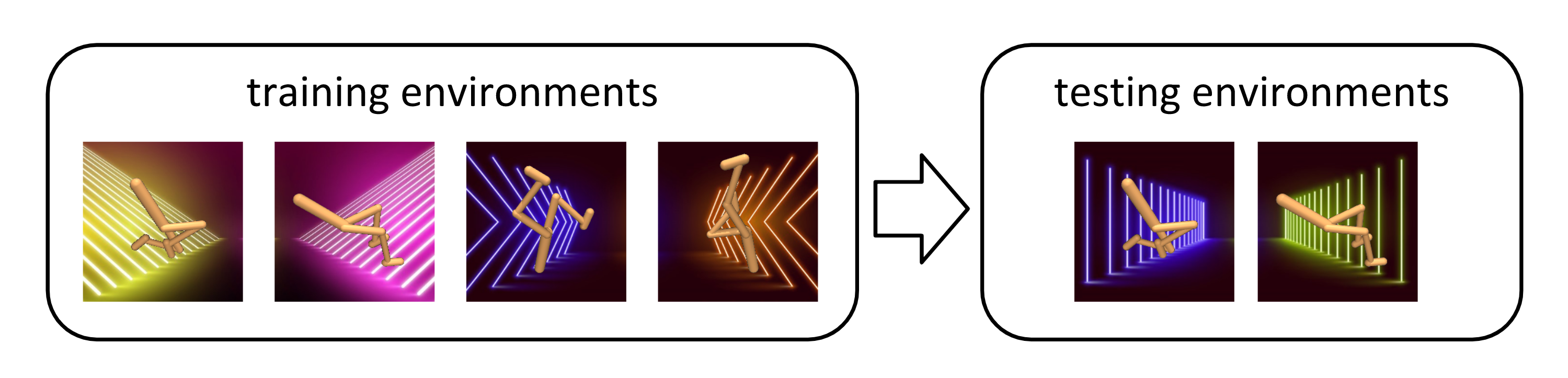}
\end{center}
\caption{Left four: training environments for the RL agent on DeepMind control tasks. Right two: testing environments in which we evaluate the RL agent. Both sets of environments use stationary backgrounds of images, the backgrounds differ simply by shape of lines and colors. 
}\label{env_images}
\end{figure*}

In this section, we empirically examine the effectiveness of the proposed method. 
In \ref{baseline}, we evaluate our method along with state-of-the-art RL methods on four continuous control tasks. 
We then examine properties of the latent space learned in \ref{sectiontsne}.
Finally, we evaluate how the trained agents generalizes zero-shot to non-stationary natural video environments.

% \thang{in general for caption, try to make it short: just the take home msg}
\subsection{Comparison against baseline}
\label{baseline}
% \textbf{TO DO: add DBC?}

Our experiments are designed to examine the following research questions: 
(1) How well do deep RL algorithms generalize to unseen states with the same underlying structure? 
(2) How does the proposed auxiliary adversarial objective compare with state of the art methods?

To answer these questions, we set up our experiments as follows: we experiment with four tasks from DeepMind Control Suite \citet{dmc} and measure zero shot generalization to unseen environments with identical task dynamics. We train the agent on four training environments, each with a different stationary background. We test the agent on two testing environments with different backgrounds. The training environments and the testing environments share identical underlying dynamics and differ simply in background visualizations. 

% \begin{figure*}[ht!]
% \begin{center}
% \includegraphics[width=0.7\textwidth]{img/illustrations2.png}
% \end{center}
% \caption{Left four: training environments for the RL agent on deepmind control tasks. Right two: testing environments in which we evaluate the RL agent. Both sets of environments use stationary backgrounds of images, the backgrounds differ simply by shape of lines and colors. 
% }\label{env_images}
% \end{figure*}

% ADD more aout baselines?

Robustness to observation changes, such as backgrounds, is vital to deploying RL in the real world. We evaluate zero shot generalization of our algorithm on a fixed set of two unseen backgrounds. The backgrounds differ from the ones in the training environment by shape of lines and color. 
% During training, the training environments are sampled sequentially, although this is only for simplicity, one can also sample the environments uniformly random.
While training on the training set, we evaluate the algorithm's ability to transfer zero-shot to the testing environments. 

We evaluate against several baselines:\\
\textbf{SAC} with data augmentation (RAD, \citet{rad}), a state-of-the-art model free RL method from images, and upon which we build DARL.\\
\textbf{DrQ} \citep{drq}, a state-of-the-art model free method from images that regularizes the value function by averaging across data augmentations.\\
% appendix/discussion
% \textbf{DBC} \citep{dbc}, a recent method that aims to learn invariant representations for RL with distracting backgrounds, by matching the l2 distance between latent representations to bisimulation metrics. \\
% \textbf{IDM} \citep{idm} - Inverse Dynamic Module, was recently shown to be an effective self-supervised objective in both training and deployment (i.e. adaptation) in visual RL \citep{pad}. We compare against a SAC baseline using IDM during training but without deployment (i.e. no adaptation or further training).

Main results are shown in Figure \ref{baseline_results}. 
See Appendix for additional baselines and results.
% DARL matches or outperforms SAC on both training and testing environments across all four tasks. 
DARL suffers from a lower generalization gap between test and train as compared to the baselines. Thanks to this better generalization, it also outperforms the baselines on all four testing tasks.
The domain adversarial module improves generalization to unseen testing environment, and in some cases also improves the rewards on training environments as it encourages efficient task-relevant representations.

\begin{figure*}[ht!]
\includegraphics[width=\textwidth]{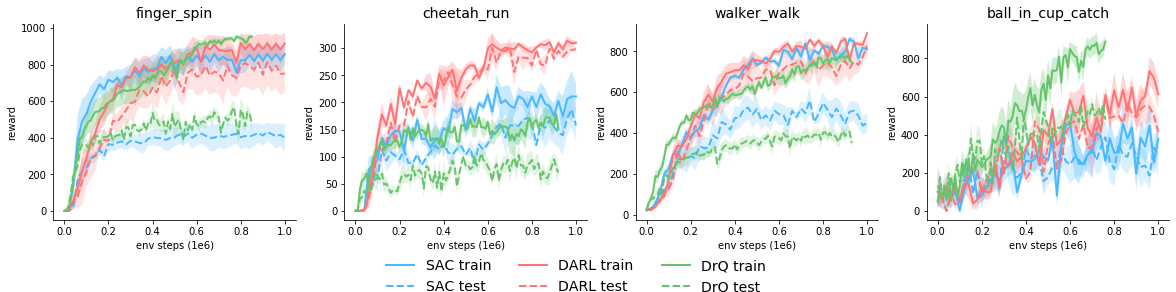}
\caption{\textbf{Deepmind control tasks.} 
% Training curves on Deepmind control.
%Joelle: --> Deepmind Control domain.
Solid and dashed lines indicate training and testing environments, respectively. 
% DARL matches or outperforms SAC on both training and testing environments across all four tasks. 
DARL outperforms SAC on testing environments across all four tasks. 
Results are averaged over 3 seeds, line shows the mean and shaded area shows standard error. 
% \textbf{DrQ is over 5 seeds, currently re-running SAC and DARL with 5 seeds}
}\label{baseline_results}
\end{figure*}

\subsection{DARL learns robust latent representations}
\label{sectiontsne}
\begin{figure}[ht!]
\includegraphics[width=0.46\textwidth]{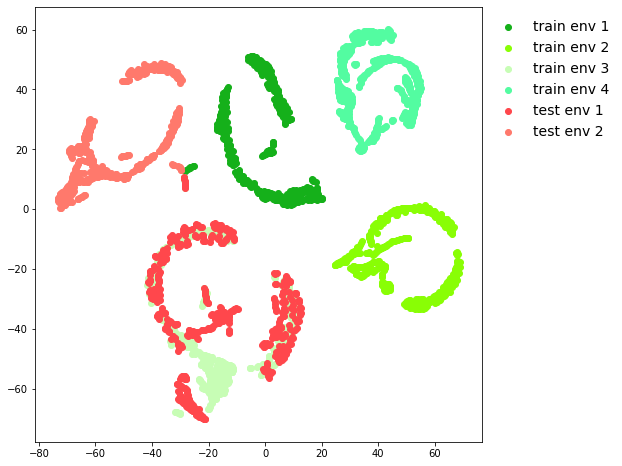}
\includegraphics[width=0.46\textwidth]{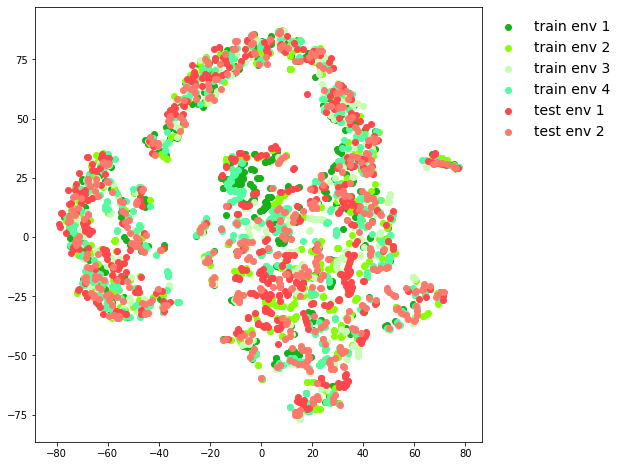}
\caption{\textbf{t-SNE of latent spaces} learned by SAC (top) and by DARL (bottom). 
Green shows state representations from training environments, red shows those from unseen testing environments. 
While SAC learns features that are background dependent,
DARL removes those task irrelevant features and learns domain-invariant representations.
}\label{tsne}
\end{figure}

To visualize the representations learned by DARL and SAC in the previous section, we use a t-SNE plot shown in Figure~\ref{tsne}. 
We collect one trajectory (one episode, 500 observations) in each training and testing environment, model the latent representations (50-dimensions) of the observations in a two-dimensional map. 
While SAC learns different embeddings for different backgrounds, our model outputs the same embedding independently of the background. This confirms that DARL could learn task relevant representations and remove features that are not necessary for the policy.

% SAC learns different representations for each background, as seen in the distinct clusters for each environment. 
% The representations learned by DARL are largely clustered together regardless of the environment the observations. 

\subsection{Zero-Shot Generalization to Non-Stationary Domains}
\label{sectionnaturalvideo}
In the previous experiment, we show that our method can generalize to test environments with different stationary backgrounds. 
We further assess generalization ability of our method by addressing the following question: 
% We now assess the generalization ability of our method by adressing the following question: 
% i think this would kinda undermine previous experiment (5.1), we have been assessing generalization, just in a different setting (stationary)
\textit{\textbf{Can DARL also zero-shot generalize to non-stationary environments?}}
% \thang{maybe quote this question?}

This setting would be more similar to real world environments where the visual setting rarely stays exactly the same. To further test the robustness of our methods, we test on environments with natural video backgrounds (\citet{natural_env}). We use the trained policies from Section \ref{baseline}, which are trained on \underline{stationary image} backgrounds, and directly evaluate them on \underline{natural video} environments.

\begin{figure*}[ht!]
\begin{center}
\includegraphics[width=0.7\textwidth]{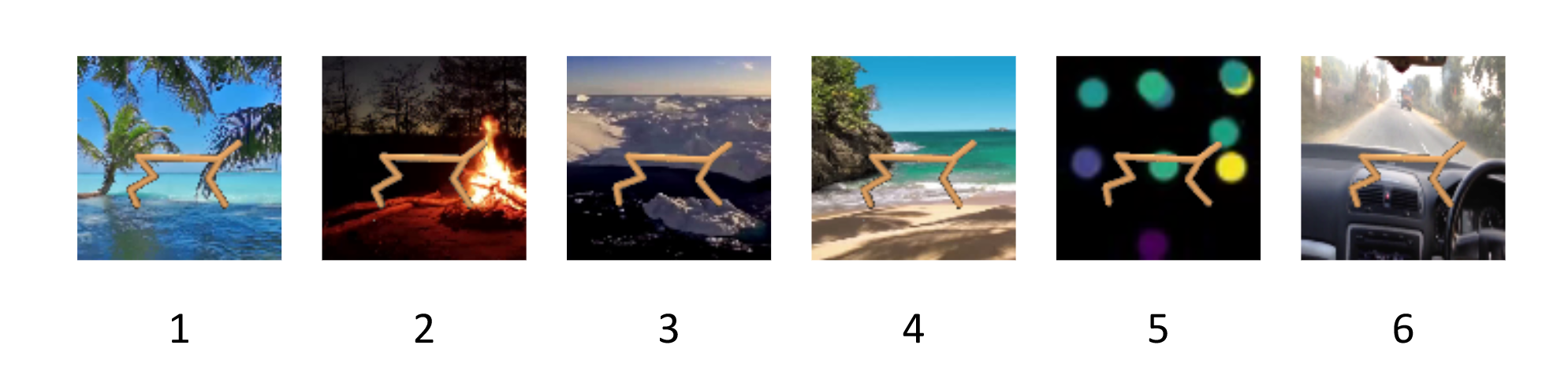}
\caption{\textbf{Non-stationary testing environments} 
The number indicates index of the environment. 
Environments all use natural video backgrounds. 
Environment 1 - 4 uses background videos from \citep{pad}; environment 5 uses simple moving distractors in the background, consisted of colored balls moving around and bouncing off the frames, as used in \citep{sac_ae}, \citep{dbc}, environment 6 uses natural car driving video from the Kinetics dataset \citep{kinetics_dataset}, as done in \citep{dbc}.
}\label{video_figure}
\end{center}
\end{figure*}

We use environments from PAD (\citet{pad}), SAC-AE (\citet{sac_ae}), and DBC (\citet{dbc}. We evaluate zero-shot generalization of the algorithms trained with the training backgrounds shown in \ref{env_images}, to natural background environments shown in Figure \ref{video_figure}. Results averaged over three environment seeds, dots indicate the mean and error bars show the standard deviation. 

To examine generalization with respect to variations of visual domains, we look at dissimilarity between such non-stationary domains. One possible measure is to use t-SNE to directly map the images into lower dimensional representations, then take the Euclidean distance between such representations. Here, we use raw frames from the natural videos, which are 30000 dimensions (3 $\times$ 100 $\times$ 100), use t-SNE directly to map them into two dimensions, compute the mean of the two dimensional representation of the natural video, and compute the L2 distance between the representation mean and that of the training domains (stationary images). We refer to this as the \textbf{video dissimilarity}, as it in general measures the visual dissimilairty between a natural video domain and the training domains. We examine the zero-shot generalization performance of our agents trained in stationary domains to video domains with respect to this dissimilarity measure.

\begin{figure*}[ht!]
\begin{center}
\includegraphics[width=0.37\textwidth]{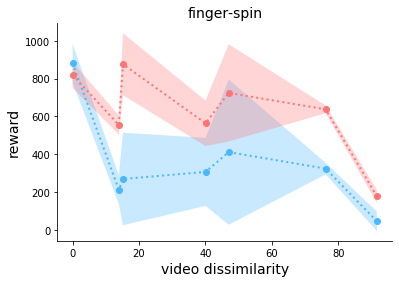}
\includegraphics[width=0.37\textwidth]{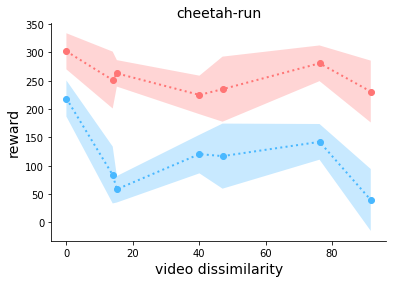}
\includegraphics[width=0.37\textwidth]{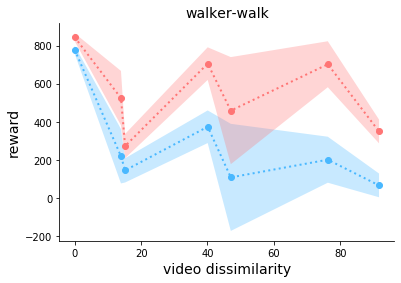}
\includegraphics[width=0.37\textwidth]{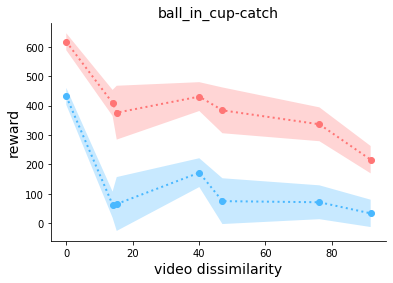}

\includegraphics[width=0.2\textwidth]{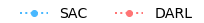}
\caption{\textbf{Generalization Rewards on natural video environments w.r.t. their image L2 distance}.
Zero-shot generalization to non-stationary environments of SAC (blue) and DARL (red). X-axis indicates the image L2 distance of the natural videos to the training images as described in \ref{sectionnaturalvideo}, 0 indicates performance on training environment with stationary images. 
}\label{video_results}
\end{center}
\end{figure*}

Results are shown in Figure \ref{video_results}. 
SAC's performance drop more drastically from the training environment.
It also varies more by environments, which we conjecture is attributable to the fact that SAC keeps information about the background in its representation, certain visual variations can lead to considerable changes in its policy. Comparatively, DARL is more robust to visual variations and more stable in performance across all environments.

\textit{\textbf{How does the generalization rewards correlate to distance in the latent representation?}}\\
% dissimilarities for visual environments
A possible measure of dissimilarities between learned representations is the distance between feature distributions in the latent space. To examine the correlation between such measure and generalization, 
% we plot the zero-shot performance with respect to the computed L2 distance of the feature mean in Figure \ref{l2_figure}.
we plot the computed L2 distance of the feature mean of the learned representations for each environment with respect to the image L2 distance of the environments.

For each natural video environment from Figure \ref{video_figure}, we compute the mean of feature distributions of its observations (averaged over 10 trajectories), and compute the L2 distance with respect to the mean of the feature distributions from the training environment. 

\begin{figure*}[ht!]
\begin{center}
\includegraphics[width=0.245\textwidth]{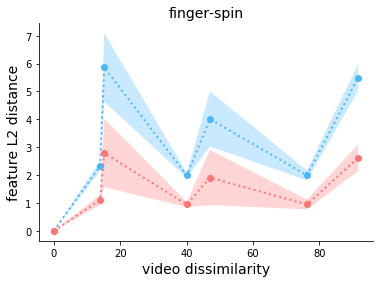}
\includegraphics[width=0.245\textwidth]{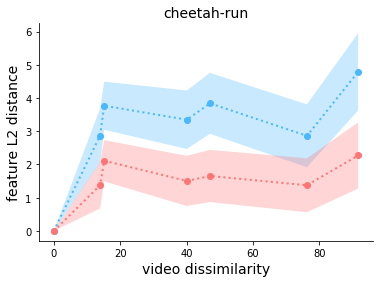}
\includegraphics[width=0.245\textwidth]{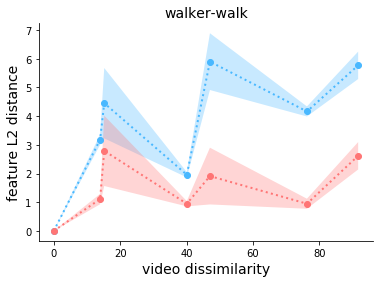}
\includegraphics[width=0.245\textwidth]{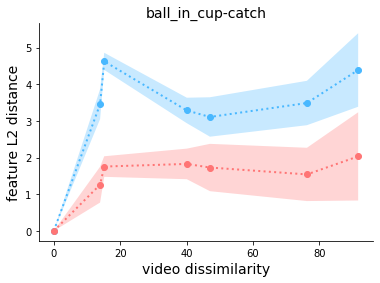}
\includegraphics[width=0.2\textwidth]{img3/legend1.png}
\end{center}
\caption{
\textbf{
 Variation of feature distributions w.r.t. dissimilarity of testing environment to training environment}, as measured by L2 distance between feature distributions of learned representations. 
The x-axis measures the t-SNE L2 distance of the natural environment from the training environment. 
As visual domains differ more from the training domains, the feature distributions learned in SAC moves further from that of the training distribution, while the distance between DARL features of the natural and training environments stays relatively small.
}\label{l2_figure}
\end{figure*}

%  Compared to DARL, SAC’s performance decreases drastically as the observation features move further from that of the training environments.
% TO BE UPDATED TO BELOW
% Variation of generalization rewards w.r.t. different environment background.} 
% As test backgrounds differs from the training backgrounds (right), baselines' accuracy is dropping drastically while DARL can maintain a decent reward.  

% We observe that generally larger distance between the feature mean corresponds to worse performance. 
% We observe a more moderate drop in generalization of DARL, and that 
Compared to SAC, DARL aligns these feature distributions from testing environments closer to that of the training environment. 
Furthermore, the drop in performance of DARL is more moderate and stable than that of SAC. On cheetah run, SAC's performance dropped by $74.1 \%$ at the testing environment furthest in latent space, while DARL's performance dropped $23.5 \%$ in its furthest testing environment. 
% Note that since DARL and SAC each has its own learned feature space and respective mappings, the numerical L2 distance in such spaces are not directly comparable. But 
Generally, as observed in Figure \ref{l2_figure}, smaller distance between the feature mean corresponds to better performance.

\section{Conclusion}
In this paper we present Domain Adversarial Reinforcement Learning (DARL), an adversarial learning approach designed to improve generalization of deep reinforcement learning agents in the context of tasks that have the same underlying dynamics but different visual observations. 
Compared to state-of-the-art model-free RL from images, DARL shows a significant improvement in zero-shot generalization to unseen domains, both stationary and non-stationary.
We hope continued research at the intersection of adversarial learning and reinforcement learning can lead to more robust deep reinforcement learning algorithms. 

\bibliography{bibliography}

\clearpage
\onecolumn
\section{Appendix}
\begin{figure*}[ht!]
\includegraphics[width=\textwidth]{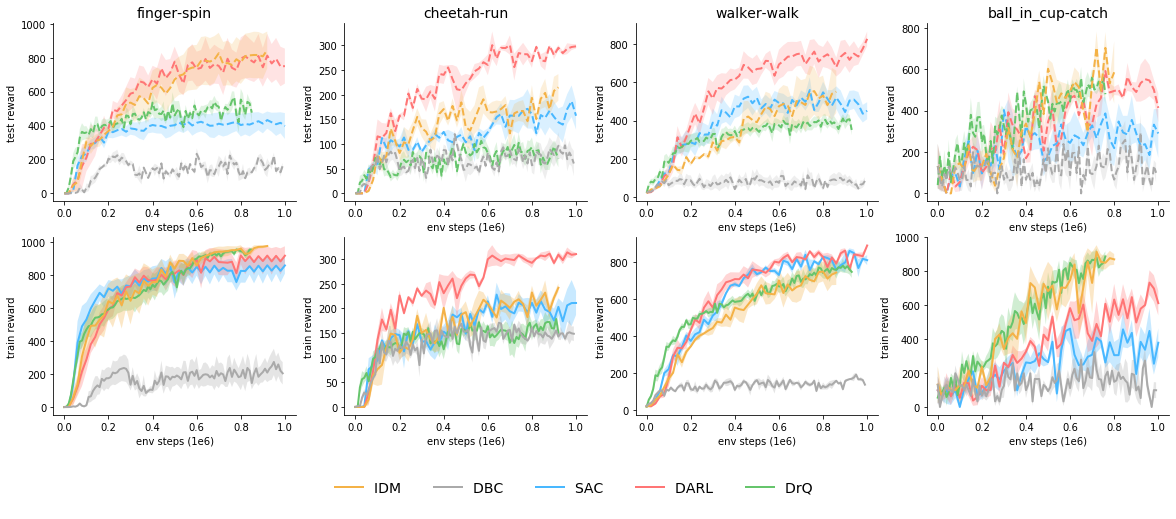}
\caption{\textbf{Baselines on DeepMind control tasks. 
Testing performance (top) and training performance (bottom).}
Line shows the mean and shaded area shows standard error, averaged across 3 seeds.
While most algorithms have good performance on training domains, they don't consistently generalize well on unseen domains; DARL generally maintains good performance across testing and training domains.
% Curves are averaged over $5$ seeds $+/-$ std.
% \textbf{DrQ is over 5 seeds, currently re-running SAC and DARL with 5 seeds}
}\label{additional_baseline}
\end{figure*}

\subsection{Additional Baselines} \label{additional_baseline_sec}
We evaluate our algorithm against two additional baselines:

\textbf{IDM} \citep{idm} - Inverse Dynamic Module 
minimizes the prediction error of the action taken given the latent representation of states and the following states. Recently, PAD \citep{pad}) uses this self-supervised objective in visual RL for both training and adaptation during policy deployment in new domains.
% uses a self-supervised objective in both training and deployment (i.e. adaptation) in visual RL (PAD, \citet{pad}). 
We compare against SAC baseline with IDM during training without adaptation, using the PAD implementation without the deployment phase. \\
\textbf{DBC} \citep{dbc} - Deep Bisimulation for Control is a recent method that aims to learn invariant representations for RL with distracting backgrounds, by matching the l2 distance between latent representations to bisimulation distances. 

We make minimal changes to standardize several experiment parameters for the baselines: environment action repeat, buffer size, and encoder architecture. Note that the original PAD encoder has 11 convolutional layers with 8 conv layers shared between the actor critic and the IDM module, which leads to slow training gain; instead we use the same encoder as SAC, DARL and DrQ with 4 convolutions layers, then share 3 layers between actor critic and the IDM. PAD originally uses a buffer size of 0.5 million, DBC uses 1 million, we use a buffer size of 100,000 in our experiments for all algorithms (following standard set up from \citet{curl, rad, drq}).

Additionally, we augment the original DBC implementation with data augmentation (random crop) used in SAC, DrQ, IDM, and DARL. We observed a small gain in DBC performance with data augmentation. However, we have yet to obtain good DBC results on some environments. We show the learning curves of DBC with data augmentation in Figure \ref{additional_baseline} for reference.

\subsection{Adversarial Objectives in DARL} \label{sec_ablation}
We investigate the set of design choices of DARL via a set of ablation experiments.
Specifically, we investigate the use of gradient reversal against another unsupervised domain adaption method. 
\textit{\textbf{Can we use a GAN-like objective for DARL?}}
ADDA \citep{adda} points that in gradient reversal the discriminator converges quickly, which leads the gradient to vanish. Instead, ADDA proposed a domain confusion objective which explicitly optimizes the discriminator output against a uniform distribution via cross-entropy loss. 
This loss would be the equivalent of GAN in the domain adaptation setting.
In our set up, the loss is
$$L_D(s, y) = - \sum_{i=1}^{n} \ln P_D(y_i \mid z), z = E_\theta(s) \;,$$
where $z$ is the feature of state $s$ from the encoder $E_\theta$,  $y_i$ indicates domain label of $i$, n is the number of training domains. 
We refer to this DARL loss variant as the \textbf{ADV loss}.

% \citet{misa} employs a variant of ADDA \citep{adda} and tries to maximize the \textbf{entropy} of the discriminator output. The loss is
% $$L_D(s, y) = \sum_{i=1}^{n} P_D(y_i \mid z) \cdot \ln P_D(y_i \mid z), z = E_\theta(s) \;,$$
% where $z$ is the feature of state $s$ from the encoder $E_\theta$, and $y_i$ indicates domain label of $i$. We refer to this as the \textbf{entropy loss}.

\begin{figure*}[ht!]
\includegraphics[width=\textwidth]{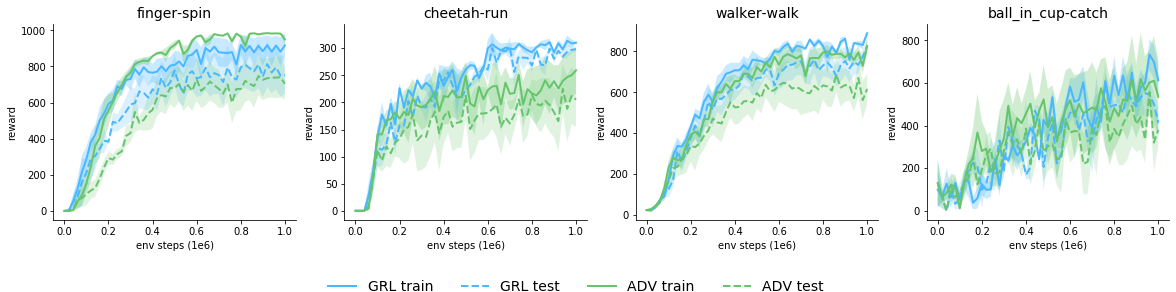}
\caption{\textbf{Adversarial Objectives in DARL.} 
Comparison between GRL and ADV. 
ADV achieves generally good performance but with sightly lower generalization reward and slightly higher variance, and requires additional run time.
% \textbf{DrQ is over 5 seeds, currently re-running SAC and DARL with 5 seeds}
}\label{gan_ablation}
\end{figure*}

We examine the gradient reversal approach that DARL uses (\textbf{GRL}) with this more GAN-like objective (\textbf{ADV}) in Figure \ref{gan_ablation}. 
While ADV also bridges the generalization gap to a certain extent, gradient reversal appears to be a generally more direct and effective method across tasks (see Figure~\ref{gan_ablation}). 

Additionally, ADV has a longer run time since it requires an additional optimization step: one backpropagation for training the discriminator without updating the encoder; and one for training the encoder using subsequent gradient from the discriminator, but without updating the discriminator. Comparatively, GRL only needs a single pass to simultaneously train the discriminator and the encoder (combined with critic loss) jointly with gradient reversal.

\subsection{Execution Time Comparison}
We examine the run time of different algorithms in table \ref{table_time}, particularly that induced by the auxiliary generalization objectives from different algorithms. 
For all algorithms, training episodes are standardized such that one training episode includes taking 500 actions (correspond to 1000 environment steps with an action repeat of 2), and one agent update after each action. 
We list one episode training time for all algorithms after 100k environment steps in table \ref{table_time}.
% (to pass the warmup time where one collects samples by acting randomly).
We list the training for the walker-walk task, though training time are very similar across all tasks.

While training time is associated with the auxiliary generalization objectives each algorithm introduces, it certainly also depends on implementation details and designs. 
For all the baselines in our experiment, we run the authors' open sourced implementation, with only minimal changes to standardize a few experiment parameters as mentioned in \ref{additional_baseline_sec}.

\begin{table*}[h!]
\centering
\begin{tabularx}{\textwidth} { 
%   | >{\centering\arraybackslash}X 
  | >{\centering\arraybackslash}X 
  | >{\centering\arraybackslash}X 
  | >{\centering\arraybackslash}X 
  | >{\centering\arraybackslash}X 
  | >{\centering\arraybackslash}X 
  | >{\centering\arraybackslash}X | }
 \hline
SAC & DrQ & IDM & DBC & DARL & DARL-adv \\
 \hline
%  finger-spin & 
 20.4 & 26.6  & 73.3  & 27.6 & 23.4 & 30.3 \\
\hline
\end{tabularx}
\caption{Execution time in seconds of algorithms for one episode iteration after 100k environment steps, averaged across 3 seeds.
DARL refers to the DARL algorithm presented in the main paper, DARL-adv refers to the DARL variant using the ADV loss described in ablation \ref{sec_ablation}}
\label{table_time}
\end{table*}

\subsection{Implementation Details}
We use the same encoder architecture as in \citet{sac_ae}. 
For all four convolutional layers, the encoder has kernel size of $3 \times 3$ with 32 channels, and uses ReLU activations; the first layer has a stride of 2, the rest has a stride of 1. 
A fully connected layer maps the flattened output of the final convolutional layer to an encoder feature dimension of 50.

The adversarial discriminator applies gradient reversal on the encoder features, uses a fully connected layer to map it to a hidden dim of 100, applies ReLU activations, before another fully connected layer mapping it to the number of training domains. We then apply LogSoftmax activations to normalize the output to log probabilities.

See Table \ref{table_param} for parameters used for the DeepMind Control experiments. See Network Architecture for printed architectures.

% \begin{table*}[h!]
% \centering
% \begin{tabularx}{0.6\textwidth} { 
%   | >{\leftslash\arraybackslash}X 
%   | >{\centering\arraybackslash}X | }
%  \hline
% Parameter name & Value \\
%  \hline
% %  finger-spin & 
%  Observation redering & $100 \times 100$
%  & Cropped observations  & $84 \times 84$
%  & Frame stack & 3
%  & Action Repeat & 2
%  & Replay buffer size & 100000
%  & Initial steps & 1000
%  & Hidden dim & 1024
%  & Optimizer & Adam
%  & Learning Rate & {2e-4 (cheetah-run) \\ \ & 1e-3 (other wise)}
%  & Critic target update frequency & 2
%  & Critic soft update rate & 0.005
%  & Critic beta  & 0.9
%  & Actor update frequency & 2
%  & Actor beta & 0.9
%  & Actor log stdev bounds & [-10, 2]
%  & Init temperature & 0.1
%  & Temperature learning rate & 1e-4
%  & Temperature beta & 0.5
%  & Evaluation episodes & 10 \\
% \hline
% \end{tabularx}
% \caption{Used hyper parameters for DMC experiments}
% \label{table_param}
% \end{table*}
% \newpage
\begin{table}[h]
    \centering
\resizebox{0.50\textwidth}{!}{
\begin{tabular}{c|c  }
\hline\hline
Parameter name & Value \\
 \hline
 Observation rendering & $100 \times 100$ \\
  Cropped observations  & $84 \times 84$ \\
  Frame stack & 3 \\
  Action Repeat & 2 \\
  Replay buffer size & 100000 \\
  Initial steps & 1000  \\
 Hidden dim & 1024  \\
  Optimizer & Adam  \\
  Learning Rate & 2e-4 (cheetah-run) \\ 
  &  1e-3 (other wise) \\ 
  Critic target update frequency & 2  \\
  Critic soft update rate & 0.005  \\
  Critic beta  & 0.9  \\
  Actor update frequency & 2  \\
  Actor beta & 0.9  \\
  Actor log stdev bounds & [-10, 2]  \\
  Init temperature & 0.1  \\
  Temperature learning rate & 1e-4   \\
  Temperature beta & 0.5  \\
  Evaluation episodes & 10 \\ \hline
\end{tabular}}
    \caption{Hyperparameters used for DMC experiments}
    \label{table_param}
\end{table}

\vspace{0.5cm}
\underline{Network Architecture}

\begin{verbatim}
Actor(
  (encoder): PixelEncoder(
    (convs): ModuleList(
      (0): Conv2d(9, 32, kernel_size=(3, 3), stride=(2, 2))
      (1): Conv2d(32, 32, kernel_size=(3, 3), stride=(1, 1))
      (2): Conv2d(32, 32, kernel_size=(3, 3), stride=(1, 1))
      (3): Conv2d(32, 32, kernel_size=(3, 3), stride=(1, 1))
    )
    (fc): Linear(in_features=39200, out_features=50, bias=True)
    (ln): LayerNorm((50,), eps=1e-05, elementwise_affine=True)
  )
(trunk): Sequential(
    (0): Linear(in_features=50, out_features=1024, bias=True)
    (1): ReLU()
    (2): Linear(in_features=1024, out_features=1024, bias=True)
    (3): ReLU()
    (4): Linear(in_features=1024, out_features=12, bias=True)
  )
)
\end{verbatim}

\newpage
\begin{spverbatim}
Critic(
  (encoder): PixelEncoder(
    (convs): ModuleList(
      (0): Conv2d(9, 32, kernel_size=(3, 3), stride=(2, 2))
      (1): Conv2d(32, 32, kernel_size=(3, 3), stride=(1, 1))
      (2): Conv2d(32, 32, kernel_size=(3, 3), stride=(1, 1))
      (3): Conv2d(32, 32, kernel_size=(3, 3), stride=(1, 1))
    )
    (fc): Linear(in_features=39200, out_features=50, bias=True)
    (ln): LayerNorm((50,), eps=1e-05, elementwise_affine=True)
  )
  (Q1): QFunction(
    (trunk): Sequential(
      (0): Linear(in_features=56, out_features=1024, bias=True)
      (1): ReLU()
      (2): Linear(in_features=1024, out_features=1024, bias=True)
      (3): ReLU()
      (4): Linear(in_features=1024, out_features=1, bias=True)
    )
  )
  (Q2): QFunction(
    (trunk): Sequential(
      (0): Linear(in_features=56, out_features=1024, bias=True)
      (1): ReLU()
      (2): Linear(in_features=1024, out_features=1024, bias=True)
      (3): ReLU()
      (4): Linear(in_features=1024, out_features=1, bias=True)
    )
  )
)
\end{spverbatim}

\begin{verbatim}
Discriminator(
  (trunk): Sequential(
    (0): Linear(in_features=50, out_features=100, bias=True)
    (1): ReLU()
    (2): Linear(in_features=100, out_features=4, bias=True)
    (3): LogSoftmax(dim=1)
  )
)
\end{verbatim}
\vspace{0.3cm}

%   (NLLLoss): NLLLoss()

\end{document}